\documentclass[conference]{IEEEtran}
\IEEEoverridecommandlockouts
\usepackage{cite}
\usepackage{amsmath,amssymb,amsfonts}
\usepackage{algorithmic}
\usepackage{graphicx}
\usepackage{textcomp}
\usepackage{xcolor}
\def\BibTeX{{\rm B\kern-.05em{\sc i\kern-.025em b}\kern-.08em
    T\kern-.1667em\lower.7ex\hbox{E}\kern-.125emX}}

\usepackage{subfig}
\usepackage [english]{babel}
\usepackage [autostyle, english = american]{csquotes}
\MakeOuterQuote{"}
\usepackage{listings}
\usepackage{enumerate} %for lists
\usepackage[section]{placeins}
\usepackage{amsmath}

\usepackage{booktabs}
\usepackage{multirow}
\usepackage{float}

\usepackage{hyperref}

\makeatletter
\let\old@ps@headings\ps@headings
\let\old@ps@IEEEtitlepagestyle\ps@IEEEtitlepagestyle
\def\confheader#1{%
\def\ps@headings{%
\old@ps@headings%
\def\@oddhead{\strut\hfill#1\hfill\strut}%
\def\@evenhead{\strut\hfill#1\hfill\strut}%
}%
\def\ps@IEEEtitlepagestyle{%
\old@ps@IEEEtitlepagestyle%
\def\@oddhead{\strut\hfill#1\hfill\strut}%
\def\@evenhead{\strut\hfill#1\hfill\strut}%
}%
\ps@headings%
}
\makeatother

\confheader{%
2019 Fifteenth International Conference on Information Processing (ICINPRO), 
December 20-22, 2019,
Bangalore, India
}

\usepackage[pscoord]{eso-pic}
\newcommand{\placetextbox}[3]{
\setbox0=\hbox{#3}
\AddToShipoutPictureFG{ \put(\LenToUnit{#1\paperwidth},\LenToUnit{#2\paperheight}){\vtop{{\null}\makebox[0pt][c]{#3}}}
}
}
\placetextbox{.2}{0.055}{  978-1-7281-3807-7/19/\$31.00~\copyright 2019 IEEE}

\begin{document}

\title{A Modified Bayesian Optimization based Hyper-Parameter Tuning Approach for Extreme Gradient Boosting}

\author{\IEEEauthorblockN{1\textsuperscript{st} Sayan Putatunda}
\IEEEauthorblockA{\textit{EDA- AA and DS CoE} \\
\textit{VMware Software India Pvt. Ltd.}\\
Bangalore, India \\
sayanp@iima.ac.in}
\and
\IEEEauthorblockN{2\textsuperscript{nd} Kiran Rama}
\IEEEauthorblockA{\textit{EDA- AA and DS CoE} \\
\textit{VMware Software India Pvt. Ltd.}\\
Bangalore, India \\
efpm04013@iiml.ac.in}

}

\maketitle

\begin{abstract}
It is already reported in the literature that the performance of a machine learning algorithm is greatly impacted by performing proper Hyper-Parameter optimization. One of the ways to perform Hyper-Parameter optimization is by manual search but that is time consuming. Some of the common approaches for performing Hyper-Parameter optimization are Grid search Random search and Bayesian optimization using Hyperopt. In this paper, we propose a brand new approach for hyperparameter improvement i.e. Randomized-Hyperopt and then tune the hyperparameters of the XGBoost i.e. the Extreme Gradient Boosting algorithm  on ten datasets by applying Random search, Randomized-Hyperopt, Hyperopt and Grid Search. The performances of each of these four techniques were compared by taking both the prediction accuracy and the execution time into consideration. We find that the Randomized-Hyperopt performs better than the other three conventional methods for hyper-paramter optimization of XGBoost.
\end{abstract}


\section{Introduction} \label{sec:introduction}
Hyper-parameter tuning plays a vital role for the optimal performance of any machine learning algorithm. Advanced machine learning algorithms such as Decision trees, Random forests, extreme gradient boosting (XGBoost), deep neural networks and support vector machines (SVM) comprises different types of hyperparameters and their tuning directly impacts the performance of the algorithm. Some algorithms such as XGBoost and Deep neural networks have many hyperparameters and this makes the tuning of hyperparameters even more challenging.  \cite{hyper:3} and \cite{hyper:21} have shown empirically that hyperparameter tuning improves the performance of the algorithm.

The Hyperparameter optimization problem can be defined as
\begin{equation}
\gamma^{(*)}=argmin_{\gamma \in \Delta} \Phi(\gamma)=  argmin_{\gamma\in \{\gamma^{(1) }\dots \gamma^{(S) } \} } \Phi(\gamma)	
\end{equation}
where, $\Phi$ is the hyperparameter response function (a.k.a. response surface), $\gamma$ represents the hyper-parameters, $\Delta$ is the search space and $\{\gamma^{(1) }\dots \gamma^{(S) } \}$ represents the trial points. Hyperparameter optimizations is defined as the minimizations of  $\Phi(\gamma)$ over $\gamma \in \Delta$ \cite{hyper:1}.

The manual optimization of hyperparameters is a simple and a well-known approach but this approach doesn't scale up especially when there are multiple settings and possibilities. The other approaches reported in the literature are that we will discuss in detail in Section \ref{lit} are Grid search, Random search and Bayesian optimization. In this paper, we will implement these methods along with a new approach for hyperparameter optimization for the Extreme gradient boosting (XGBoost) method. 

The reason for the choice of XGBoost for our experiments in this paper is that for large scale machine learning problems on structured datasets, XGBoost is a very popular method and is used quite often for solving different business problems. According to the machine learning literature, a properly tuned XGBoost always performs better in terms of prediction accuracy than that of a not properly tuned XGBoost. Thus development of new solutions for improving the accuracy of XGBoost is a very important and relevant contribution to the computational management science literature.

In this paper, we propose a  new approach for hyperparameter optimization i.e. Randomized-Hyperopt.  This technique is a variant of the Bayesian approximation method using Hyperopt (i.e. a Python library). We then tune the hyperparameters of the XGBoost i.e. the Extreme Gradient Boosting algorithm  on ten datasets by applying Random search, Randomized-Hyperopt, Hyperopt and Grid Search. The performances of each of these four techniques were compared by taking into account both the prediction accuracy and the execution time. We find that the hyperparameter optimization of XGBoost using Randomized-Hyperopt is the best performer. To the best of our knowledge, this has not been reported in the literature earlier. Some of the contributions of this study are in the development of a new approach for hyper-parameter optimization of an advance machine learning algorithm such as XGBoost and also in performing a comparative study of the performance of our proposed method with other traditional approaches for hyperparameter optimization of XGBoost. 

The remainder of this paper is structured as follows. Section \ref{lit} gives a brief review of literature of the work done in this field. In Section \ref{data} we discuss the various datasets used for our experiments. It is then followed by Section \ref{method} where we discuss the methodology. Then in Section \ref{exp} we discuss the implementation, evaluation metrics and the results of all the experiments conducted. Finally, we conclude this paper in Section \ref{con}.

\section{Background and Related Work}\label{lit}
Grid search is a very popular and a widely used method for Hyperparameter optimization where it is used to search through a manually defined subset of hyperparameters of a machine learning algorithm \cite{hyper:2}. Moreover, Grid search is easy to implement and performs parallel operations easily. However, performing Grid search turns out to be computationally expensive with the increase in hyperparameters \cite{hyper:2}.

An alternative to the Grid search method for hyperparameter optimization is the Random search method where a generative process is used to draw random samples. This generative process is used for defining the configuration space, drawing the assignments for hyperparameters and then they are  evaluated \cite{hyper:3}. Random search works more efficiently than Grid search in high dimensional spaces and generally it is found that Random search performs better than Grid search in most cases \cite{hyper:1}.

Another powerful method for hyperparameter optimization reported in the literature is the Bayesian optimization. \cite{hyper:4}  describes Bayesian optimization as a "black box" technique. \cite{hyper:7} explains the modus-operandi of the Bayesian optimization method. First, a prior measure over the objective space is chosen and one of the popular choice for the prior function is the Gaussian Process (GP). Then we get the posterior measure given some observation over the objective function by combining the prior and the likelihood. Finally, the next evaluation is computed by taking into account the loss function. Please refer to \cite{hyper:7} for more details on Bayesian optimization. This approach i.e. Bayesian optimization for hyperparameter tuning is also known as "Sequential Model-Based Global Optimization (SMBO)" \cite{hyper:3,hyper:5}.

The SMBO algorithm is computationally very effective with costly fitness functions. The way it works is that the  expensive fitness function say $g$ is approximated by using a cheaper to evaluate surrogate function \cite{hyper:3}. The goal of the SMBO method is to reach the point  $x^{*}$ where the surrogate function is maximized and then this point  $x^{*}$ is proposed to evaluate the tyrue function $g$ \cite{hyper:3}. SMBO can handle high dimensional data and perform parallel evaluations effectively \cite{hyper:20}. 

%\begin{figure}[!htp]
%\centering
%\includegraphics[width=0.45\textwidth]{Fig1.eps}
%\caption{SMBO process flow}
%\label{fig:smbo}
%\end{figure}
%
%Figure \ref{fig:smbo} shows the different steps or stages in executing the SMBO algorithm. 
%Instead of the Gaussian Process (GP) where the objective is to model $p(y|x))$, one can also go for the Tree-structured Parzen Estimator (TPE) approach that models $p(y)$ and  $p(x|y)$ \cite{hyper:3}. The TPE method uses non-parametric densities to model $p(x|y)$ and uses a variation of gaussian mixture to replace prior distributions such as uniform and log-uniform \cite{hyper:3}.

SMBO algorithm can be implemented and applied on various machine learning algorithms for hyperparameter optimization using the Python library known as "Hyperopt" (see Section \ref{method} for more details). \cite{hyper:20} used Hyperopt for hyperparameter optimization of convolutional neural networks and deep neural networks. Some studies reported in the literature have applied Hyperopt on other advanced machine learning algorithms such as Random forests and Support vector machines \cite{hyper:23}.  

Hyperopt was primarily developed for research on Deep learning \cite{hyper:3} and Computer vison \cite{hyper:22}. \cite{spml:1} worked on comparing the performance of Hyperopt with respect to the Grid search and Random search for hyperparameter optimization of XGBoost. They found that Hyperopt performs consistently better than the other two approaches in terms of prediction accuracy. In this paper we extend the work by proposing a new approach for hyperparameter tuning of Extreme Gradient Boosting and then compare its performance with that of conventional techniques namely, Grid search, Random search and Bayesian optimization using Hyperopt. Moreover, in this paper we experiment with more datasets (please see Section \ref{data} for more details).

The three approaches  mentioned above for hyperparameter optimization viz.  Bayesian optimization, Grid search and Random search are the most popular and widely used in the academia and the industry. However, there have been studies reported in the literature that proposed other alternative approaches for hyperparameter optimization of machine learning algorithms.

Li et al. \cite{hyper:11} proposed a technique called Hyperband, which is basically a multi-arm bandit strategy that dynamically allocates resources to configurations that are randomly sampled and this is based on the performance of these randomly sampled configurations on subsets of the data. Hyperband discards the poorly-performing configurations early and ensures that only the better performing configurations are trained on the entire dataset.

Also, a number of optimization methods have been used for hyperparameter search as mentioned in the literature. Some of the notable ones are genetic algorithms \cite{hyper:12}, coupled simulated annealing \cite{hyper:16}, racing algorithms \cite{hyper:15} and swarm optimization \cite{hyper:13,hyper:14}.  \cite{hyper:17} compared several Artificial Neural Networks (ANN) weight initialization using Evolutionary Algorithms (EA). Some studies \cite{hyper:18,hyper:19} have shown that in hyperparameter tuning, Evolutionary Algorithms perform better than Grid search techniques when the accuracy-speed ratio is considered.

\section{Datasets} \label{data}
In this paper, ten publicly available datasets were taken from the UCI machine learning repository \cite{uci:1}. Table \ref{data:des} describes the number of attributes and observations for each of the ten datasets. Some of these benchmark datasets were also used in many studies viz. \cite{data:2}, \cite{spml:1}, \cite{data:4} and others. The relevant details for each datasets are discussed below. 

\textsc{banknote:} This dataset comprises of data extracted from images that were taken for the evaluation of an authentication procedure of banknotes. The features were extracted from the images using the Wavelet transform tool. The target variable indicates if the banhnotes are authentic and the predictor variables are features from the images such as variance, curtosis and skewness of the wavelet transformed image along with the entropy of the image \cite{uci:1}. 

\textsc{contraceptive:}  The dependent variable is converted to a binary variable that indicates whether a woman is using contraceptive for short/long term or not using it at all. The various predictor variables are the socio-economic and demographic characteristics of each women. This dataset was also used in \cite{data:8}. 

\textsc{transfusion:}  This dataset comprises the data for donor taken from a blood transfusion service center in Taiwan. Here, the depeendent variable indicates if the donor donates blood \cite{data:9}. 

\textsc{acceptability:} The dependent variable is a binary variable indicates the acceptability of a car. The various features include the price, comfort and technical characteristics of the cars. This dataset was first used in \cite{data:10}. 

\textsc{subscribe:} In this dataset, the dependent variable represents whether a customer has a term deposit subscription or not. The different features include the  different demographic and socio-economic characteristics of the customers along with details regarding loan taken and credit default \cite{data:2}. 

\textsc{retinopathy:} This dataset had features are extracted from the Messidor image datset.  The dependent variable indicates the presence of Diabetic retinopathy for a particular patient  \cite{data:3}.  

\textsc{credit-default:}  The different predictor variables were the demographic and payment information of the customers. The dependent variable represents default by the concerned bank's customers \cite{data:4}.

\textsc{regularity:}  In this paper we are using a modified dependent variable (from the absenteeism at workplace dataset  \cite{data:5}) that indicates whether a person is regular at work or not. The different predictor variables include age, drinking habits, service time, work load and more. 

\textsc{income:}  It is also known as the Census income dataset for the census conducted in the year 1994 \cite{data:7}. The target variable is whether the income exceeds 50k USD per year and the different predictors are age, workclass, education, occupation, gender, race, country and other information of the individuals. 

 \textsc{recurrence:} This dataset  was first used in \cite{data:6} and the dependent variable indicates the recurrence/non-recurrence of cancer. 

\begin{table}[!htp]
\centering
\caption{Dataset Description}
\label{data:des}
  \scalebox{0.7}{
\begin{tabular}{lcc}
\\ \hline  
Dataset      & no. of attributes & no. of observations \\ \hline  
BANKNOTE      & 5                    & 1372                   \\\hline  
CONTRACEPTIVE & 9                    & 1473                   \\\hline  
TRANSFUSION   & 5                    & 748           \\ \hline     
ACCEPTABILITY          & 6                    & 1728                   \\\hline  
SUBSCRIBE          & 20                   & 45211                  \\\hline  
RETINOPATHY      & 20                   & 1151                   \\\hline  
CREDIT-DEFAULT       & 24                   & 30000                  \\\hline  
REGULARITY        & 21                   & 740                    \\\hline  
INCOME        & 14                   & 48842                  \\\hline  
RECURRENCE        & 9                    & 286                    \\\hline       
\end{tabular}}
\end{table}

\section{Methodology} \label{method}
In Section \ref{lit}, we have described Hyperopt as a Python library that implements the SMBO algorithm for hyperparameter optimization of machine learning algorithms. The way Hyperopt operates is that first we define the configuration space using different types of distributions (namely, normal/uniform/log-uniform). The use of "quantized" continuous distributions such as qnormal, qlognormal and quniform are also allowed. Please see \cite{hyper:20} for a detailed discussion on different types of distributions that can be used in Hyperopt to define the configuration space. The next step is to then optimize using the \textit{fmin} driver.

\begin{figure}[!htp]
\centering
\includegraphics[width=0.45\textwidth]{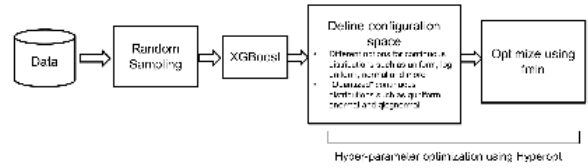}
\caption{Randomized-Hyperopt Framework}
\label{fig:random}
\end{figure}

In this paper we propose the Randomized-Hyperopt method, which is a variant of the bayesian optimization approach for hyperparameter tuning using Hyperopt. Figure \ref{fig:random} shows the Randomized-Hyperopt solution framework. Here instead of directly using the entire data for classification task using XGBoost on which Hyperopt is applied to generate the optimal hyperparameters, we first perform a random sampling. The idea is that since a sample represents the entire population so we can use a sample instead of the entire training data and then proceed with applying Hyperopt to generate the optimal hyperparameters of the XGBoost algorithm as shown in Figure \ref{fig:random}. We can use these hyperparameters to perform prediction on unseen data and evaluate the prediction accuracy. This approach reduces the execution time to generate the optimal hyperparameters drastically when compared to the other methods (see Table \ref{compare} for more details). We applied this approach on $10$ benchmark datasets (see Section \ref{data}) and compared it with other existing techniques for hyperparameter optimization. Randomized-Hyperopt is a totally new approach for hyperparameter optimization and it hasn't been reported in the literature earlier.

\section{Experimental Results} \label{exp}
All the experiments were performed on a system with configurations 4 GB RAM  1.6 GHz Intel Core i5 Mac OSX and using the open-source software Python 3 (with libraries such as hyperopt and scikit-learn).

\begin{figure}[!htp]
\centering
\includegraphics[width=0.5\textwidth]{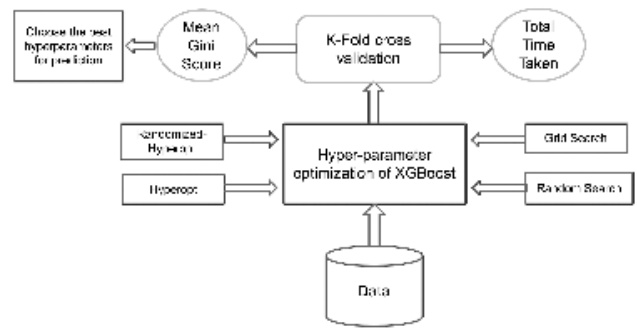}
\caption{The Experimental Design}
\label{fig:design}
\end{figure}

Figure \ref{fig:design} shows our experimental design. For all our experiments with Randomized-Hyperopt and Hyperopt, the search space for the different hyperparameters of XGBoost are defined first. These hyperparametrs are defined within a range of values. We perform Randomized-Hyperopt with different sampling rates ($r\%$)  and then select the best performing one. While executing the experiments, a Stratified K-fold cross validation is used (where, $K=3$).The mean Gini scores are computed for each of the four techniques (see Section \ref{eval}) to reach the optimum value of the  hyperparameters. Similar experiments are performed for Grid search and Random search as well. We use the GridSearchCV and RandomizedSearchCV  functions of the Scikit-learn library with $10$ iterations for the same.

In Section \ref{eval}, we first discuss the evaluation metrics used in this paper which is followed by the Results of all the experiments conducted in this study in Section \ref{res}.

\subsection{Evaluation Metric} \label{eval}
The evaluation metric used in this study is the "Mean Gini Score". In the economics and the sociology literature, the Gini score is a very popular metric because of its clear economic interpretation \cite{gini:3}. The Lorentz curve \cite{lorentz:1} is used for analysis in various scientific problems and the Gini score is a summary index in it \cite{gini:2}. Generally, in classification problems, the ROC (Receiver Operating Characteristics) Area under the curve (AUC) iscommonly used as a metric that signifies prediction accuracy of the classification algorithm. However, there is a clear connection between Lorentz curve/Gini score and ROC AUC as shown in \ref{equ:1} below. This was first discussed in \cite{gini:1}.

\begin{equation}
\label{equ:1}
G=2*ROC\_AUC-1
\end{equation}
where $ROC\_AUC$= ROC Area under the curve and G= Gini Score.
As mentioned earlier, the Stratified K-fold cross validation is applied and the mean gini score is computed for each of the methods. The higher the mean Gini score, the better is the performance in terms of prediction accuracy. So while comparing, a method with the highest mean Gini score is the best performer. We also keep a track of the execution time (in seconds) of each of the four methods for generating the best parameters.

\subsection{Results}\label{res}
In Table \ref{samples} the mean Gini score is mentioned along with the time taken (in Seconds) for Randomized-Hyperopt to generate the best parameters with different sampling rates ($r\%$) viz. $10\%$, $20\%$, $25\%$ and $50\%$ for the Extreme gradient boosting algorithm on each of the ten datasets. Then we chose the best sampling rate for Randomized-Hyperopt for each of the datasets using both mean Gini score and the time taken. For the rest of this paper, we choose the sampling rate of Randomized-Hyperopt to be (a) $10\%$ for the \textsc{subscribe} dataset, (b) $20\%$ for the \textsc{acceptability}, \textsc{retinopathy}and \textsc{banknote} datasets, (c) $25\%$ for the \textsc{income} dataset and (d) $50\%$ for the remaining datasets viz. \textsc{credit-default}, \textsc{regularity}, \textsc{recurrence}, \textsc{contraceptive} and \textsc{transfusion}.

\begin{table}[!htp]
\centering
\caption{Randomized-Hyperopt with different Sampling rates for each datasets}
\label{samples}
 \scalebox{0.7}{
\begin{tabular}{lccc}
\hline
Dataset                        & Sampling rate & Mean Gini & Time (in Seconds) \\ \hline
\multirow{4}{*}{BANKNOTE}      & 10\%          & 0.9993    & 0.22        \\
                               & 20\%          & 0.9998    & 0.31        \\
                               & 25\%          & 0.9998    & 0.40         \\
                               & 50\%          & 0.9998    & 0.59        \\ \hline
\multirow{4}{*}{CONTRACEPTIVE} & 10\%          & 0.4457    & 0.32        \\
                               & 20\%          & 0.4567    & 0.43        \\
                               & 25\%          & 0.4416    & 0.58        \\
                               & 50\%          & 0.4644    & 0.82        \\ \hline
\multirow{4}{*}{TRANSFUSION}   & 10\%          & 0.2632    & 0.21        \\
                               & 20\%          & 0.2858    & 0.30         \\
                               & 25\%          & 0.3473    & 0.42        \\
                               & 50\%          & 0.3556    & 0.52       \\ \hline
\multirow{4}{*}{ACCEPTABILITY}          & 10\%          & 0.8307    & 0.47        \\
                               & 20\%          & 0.8332    & 0.85        \\
                               & 25\%          & 0.8159    & 1.12        \\
                               & 50\%          & 0.825     & 2.12        \\ \hline
\multirow{4}{*}{SUBSCRIBE}          & 10\%          & 0.7597    & 1.11        \\
                               & 20\%          & 0.7308    & 2.67        \\
                               & 25\%          & 0.7539    & 3.18        \\
                               & 50\%          & 0.7243    & 8.24        \\ \hline
\multirow{4}{*}{RETINOPATHY}      & 10\%          & 0.4926    & 0.34        \\
                               & 20\%          & 0.5151    & 0.40         \\
                               & 25\%          & 0.4961    & 0.50         \\
                               & 50\%          & 0.4984    & 0.67        \\ \hline
\multirow{4}{*}{CREDIT-DEFAULT}       & 10\%          & 0.5455    & 16.27       \\
                               & 20\%          & 0.5493    & 39.82       \\
                               & 25\%          & 0.5477    & 42.67       \\
                               & 50\%          & 0.5594    & 55.90        \\ \hline
\multirow{4}{*}{REGULARITY}        & 10\%          & 0.9658    & 0.39        \\
                               & 20\%          & 0.9321    & 0.60        \\
                               & 25\%          & 0.9676    & 0.69        \\
                               & 50\%          & 0.9718    & 1.10         \\ \hline
\multirow{4}{*}{INCOME}        & 10\%          & 0.8431    & 47.69       \\
                               & 20\%          & 0.8557    & 125.55      \\
                               & 25\%          & 0.8563    & 106.39      \\
                               & 50\%          & 0.8559    & 284.49      \\ \hline
\multirow{4}{*}{RECURRENCE}        & 10\%          & 22.53     & 0.28        \\
                               & 20\%          & 24.63     & 0.29        \\
                               & 25\%          & 22.55     & 0.35        \\
                               & 50\%          & 25.93     & 0.39        \\ \hline
\end{tabular}}
\end{table}

\begin{table}[!htp]
\centering
\caption{Mean Gini and the time taken for each methods on the ten datasets}
\label{compare}
 \scalebox{0.7}{
\begin{tabular}{llcc}
\hline
Dataset                        & Method              & \multicolumn{1}{l}{Mean Gini} & \multicolumn{1}{l}{Time (in Seconds)} \\ \hline
\multirow{4}{*}{BANKNOTE}      & Randomized-Hyperopt & 0.9998                        & 0.31                            \\
                               & Hyperopt            & 0.9999                        & 1.14                            \\
                               & Random Search       & 0.9635                        & 1.21                            \\
                               & Grid Search         & 0.8806                        & 86.62                           \\ \hline
\multirow{4}{*}{CONTRACEPTIVE} & Randomized-Hyperopt & 0.4644                        & 0.82                            \\
                               & Hyperopt            & 0.4672                        & 6.00                               \\
                               & Random Search       & 0.4411                        & 3.35                            \\
                               & Grid Search         & 0.3741                        & 322.90                           \\ \hline
\multirow{4}{*}{TRANSFUSION}   & Randomized-Hyperopt & 0.3556                        & 0.52                            \\
                               & Hyperopt            & 0.3510                         & 0.85                            \\
                               & Random Search       & 0.2607                        & 0.88                            \\
                               & Grid Search         & 0.2276                        & 61.48\\   \hline
\multirow{4}{*}{ACCEPTABILITY}          & Randomized-Hyperopt & 0.8332                        & 0.85                            \\
                               & Hyperopt            & 0.8445                        & 3.35                            \\
                               & Random Search       & 0.7428                        & 3.10                            \\
                               & Grid Search         & 0.7726                        & 139.52                          \\ \hline
\multirow{4}{*}{SUBSCRIBE}          & Randomized-Hyperopt & 0.7597                        & 1.11                            \\
                               & Hyperopt            & 0.7518                        & 28.30                            \\
                               & Random Search       & 0.7114                        & 14.13                           \\
                               & Grid Search         & 0.6742                        & 569.74                          \\ \hline
\multirow{4}{*}{RETINOPATHY}      & Randomized-Hyperopt & 0.5151                        & 0.40                             \\
                               & Hyperopt            & 0.5055                        & 5.45                            \\
                               & Random Search       & 0.4906                        & 2.87                            \\
                               & Grid Search         & 0.4707                        & 289.98                          \\ \hline
\multirow{4}{*}{CREDIT-DEFAULT}       & Randomized-Hyperopt & 0.5594                        & 55.90                            \\
                               & Hyperopt            & 0.5467                        & 180.29                          \\
                               & Random Search       & 0.5003                        & 115.88                          \\
                               & Grid Search         & 0.5258                        & 3586.23                         \\ \hline
\multirow{4}{*}{REGULARITY}        & Randomized-Hyperopt & 0.9718                        & 1.10                             \\
                               & Hyperopt            & 0.9739                        & 2.09                            \\
                               & Random Search       & 0.8710                         & 1.68                            \\
                               & Grid Search         & 0.8593                        & 141.15                          \\ \hline
\multirow{4}{*}{INCOME}        & Randomized-Hyperopt & 0.8563                        & 106.39                          \\
                               & Hyperopt            & 0.8545                        & 647.81                          \\
                               & Random Search       & 0.8246                        & 265.65                          \\
                               & Grid Search         & 0.8155                        & 84879.56                        \\ \hline
\multirow{4}{*}{RECURRENCE}        & Randomized-Hyperopt & 0.2593                         & 0.39                            \\
                               & Hyperopt            & 0.2497                        & 1.70                             \\
                               & Random Search       & 0.1961                        & 1.05                            \\
                               & Grid Search         & 0.1670                         & 98.54                           \\ \hline
\end{tabular}}
\end{table}

In Table \ref{compare}, we compare the performance and the time taken to identify the best parameters for each of the four hyperparameter optimization techniques viz. Randomized-Hyperopt,  Grid search, Hyperopt, and Random search for each of the ten datasets. We find that in terms of mean Gini the proposed Randomized-Hyperopt method performs better than Grid search and Random search methods for all the datasets consistently and it is either close to or greater than the mean Gini of the Hyperopt method for all the datasets. For the \textsc{transfusion}, \textsc{subscribe}, \textsc{retinopathy}, \textsc{credit-default}, \textsc{income} and \textsc{recurrence} datasets the mean Gini of Randomized-Hyperopt is greater than that of Hyperopt whereas for the remaining datasets, the mean Gini of Randomized-Hyperopt is very close to that of the Hyperopt method. However, when we take the the time taken into consideration then the Randomized-Hyperopt method is the fastest consistently across all the ten datasets followed by the Random search method. Thus, taking into account both the time and the prediction accuracy, Randomized-Hyperopt is the recommended method for the hyperparameter tunning of the Extreme gradient boosting algorithm.

\section{Conclusion} \label{con}
In this paper, we propose a new method for hyperparameter optimization of the Extreme gradient boosting algorithm i.e. Randomized-Hyperopt and compare its performance (by taking both the prediction accuracy and the time taken into consideration) with other existing techniques such as Grid search, Random search and Bayesian optimization using Hyperopt. We find that the Randomized-Hyperopt method performs better than the bayesian optimization using Hyperopt, Grid search and Random search methods for all the datasets consistently and it is either close to or greater than the mean Gini of the Hyperopt method for all the datasets. Also, the Randomized-Hyperopt method takes the least execution time consistently across all the ten datasets. Thus, taking into account both the time and the prediction accuracy, Randomized-Hyperopt is the recommended method for the hyperparameter optimization of the  Extreme gradient boosting algorithm.

As a direction for future research, we would like to explore how Randomized-Hyperopt performs for hyperparameter optimization of other advanced machine learning algorithms such as Deep neural networks, Random forests and Support vector machines. In this paper, we have focused on classification problem but it will be interesting to see how Randomized-hyperopt performs in a regression setting. Also, we would like to explore how Randomized-Hyperopt performs compared to other techniques for hyperparameter optimization that uses techniques such as Genetic algorithms and Swarm optimization.

\bibliographystyle{IEEEtran}
\bibliography{mybib} 
\end{document}